\documentclass[10pt,twocolumn,letterpaper]{article}

\usepackage{3dv}
\usepackage{times}
\usepackage{epsfig}
\usepackage{graphicx}
\usepackage{amsmath}
\usepackage{amssymb}
\usepackage{bbm}
\usepackage{tabularx}
\usepackage{multirow}
\usepackage{multicol}
\usepackage{subcaption}
\usepackage{rotating}
\usepackage{gensymb}
\usepackage[font=small]{caption}
\usepackage{microtype}
\usepackage{enumitem}
\usepackage{authblk}

\usepackage[pagebackref=true,breaklinks=true,letterpaper=true,colorlinks,bookmarks=false]{hyperref}

\def\*#1{\mathbf{#1}}
\def\rs#1{r\rightarrow s_{#1}}
\def\sr#1{s_{#1}\rightarrow r}
\def\ab{a \rightarrow b}
\def\ba{b \rightarrow a}

\threedvfinalcopy 
\def\threedvPaperID{304}

\ifthreedvfinal\fi

\newcommand{\ignore}[1]{}

\newcommand\vrulel[1]{%
  \multicolumn{1}{|c}{#1}%
}

\newcommand\vruler[1]{%
  \multicolumn{1}{c|}{#1}%
}

\begin{document}

\title{Deep Multi-View Stereo Gone Wild}

\author[1,2]{Fran\c{c}ois Darmon}
\author[1]{Bénédicte Bascle}
\author[1]{Jean-Clément Devaux}
\author[2]{Pascal Monasse}
\author[2]{Mathieu Aubry}
\affil[1]{\small Thales LAS France}
\affil[2]{\small LIGM (UMR 8049), \'Ecole des Ponts, Univ. Gustave Eiffel, CNRS, Marne-la-Vall\'ee, France}
\affil[ ]{\url{http://imagine.enpc.fr/\~darmonf/wild_deep_mvs/}}

\maketitle

\ifthreedvfinal\thispagestyle{empty}\fi

\begin{abstract}
Deep multi-view stereo (MVS) methods have been developed and extensively compared on simple datasets, where they now outperform classical approaches. In this paper, we ask whether the conclusions reached 
in controlled scenarios are still valid when working with Internet photo collections. We propose a methodology for evaluation and explore the influence of three aspects of deep MVS methods: network architecture, training data, and supervision. We make several key observations, which we extensively validate quantitatively and qualitatively, both for depth prediction and complete 3D reconstructions. 
First, complex unsupervised approaches cannot train on data in the wild. Our new approach makes it possible with three key elements: upsampling the output, softmin based aggregation and a single reconstruction loss. Second, supervised deep depthmap-based MVS methods are state-of-the art for reconstruction of few internet images. Finally, our evaluation provides very different results than usual ones. This shows that evaluation in uncontrolled scenarios is important for new architectures.  

\end{abstract}

\section{Introduction}

\begin{figure}
    \centering
    \tabcolsep=0.05cm
    \begin{tabular}{cccc}
        \rotatebox[origin=c]{90}{\parbox{2cm}{\centering DTU\\ (lab settings)}} & 
        \begin{subfigure}{0.55\columnwidth}
            \includegraphics[height=47pt]{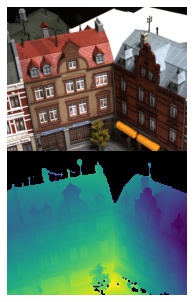}
             \includegraphics[height=47pt]{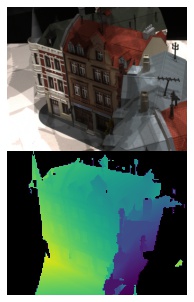}
              \includegraphics[height=47pt]{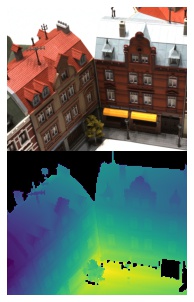}
               \includegraphics[height=47pt]{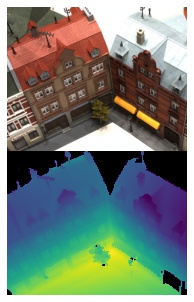}       
        \end{subfigure} & 
        $\boldsymbol{\cdots}$ &
        \begin{subfigure}{0.3 \columnwidth}
                \includegraphics[width=\linewidth,trim={0, 100pt, 300pt, 0}, clip]{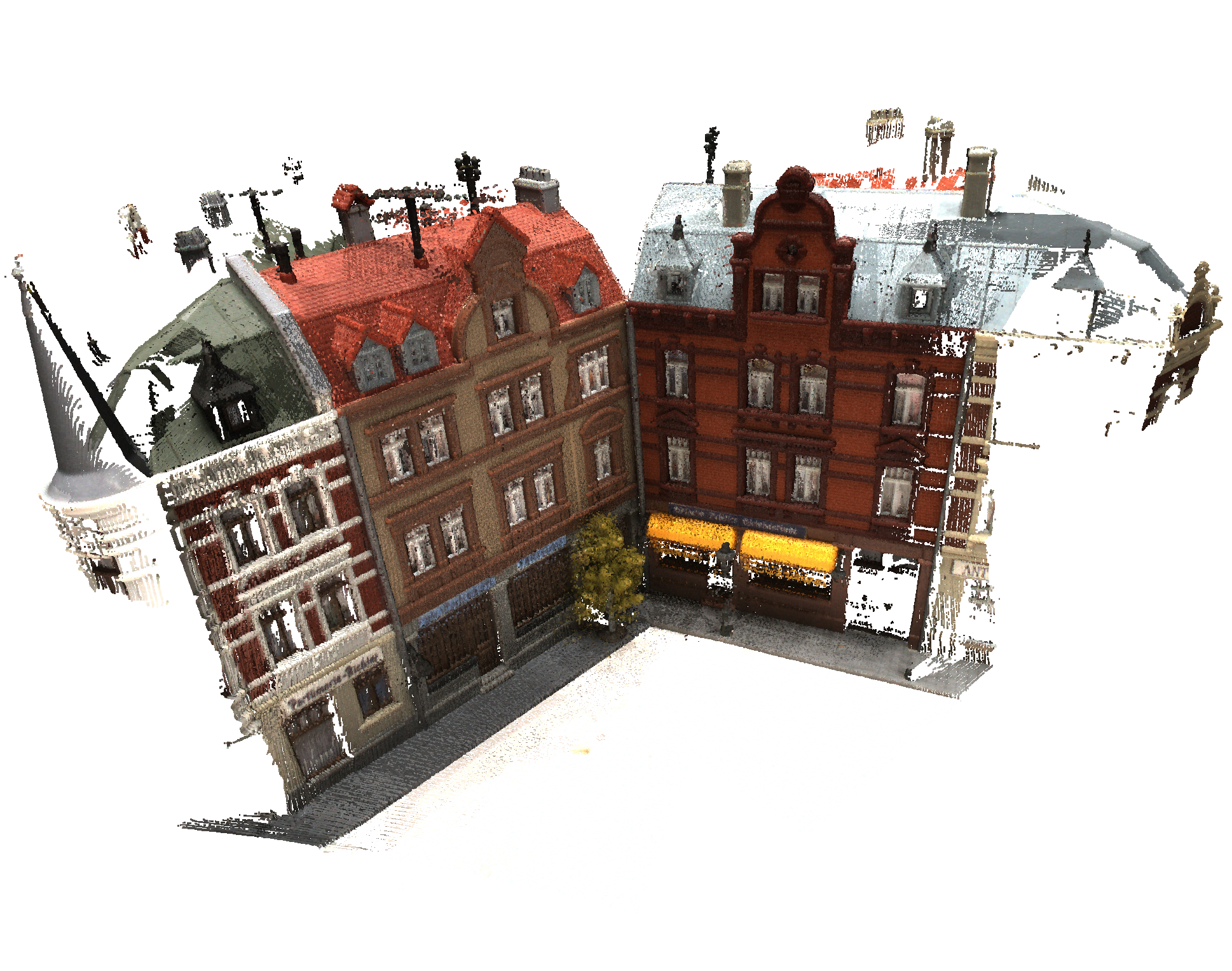}
        \end{subfigure}        
        \vspace{-3mm}
        \\
        \rotatebox[origin=c]{90}{\parbox{3cm}{\centering Blended MVS \\(synthetic)}} & 
        \begin{subfigure}{0.55\columnwidth}
            \centering
            \includegraphics[height=64pt]{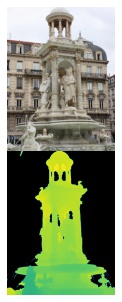}
            \hfill
            \includegraphics[height=64pt]{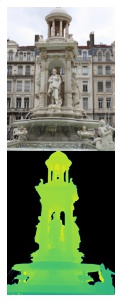}
            \hfill
            \includegraphics[height=64pt]{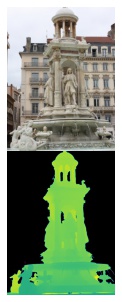}
            \hfill
            \includegraphics[height=64pt]{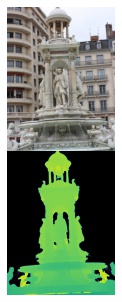}
        \end{subfigure} & 
        $\boldsymbol{\cdots}$ & 
        \begin{subfigure}{0.3 \columnwidth}
            \includegraphics[height=64pt,trim={200pt, 0, 200pt, 0}, clip]{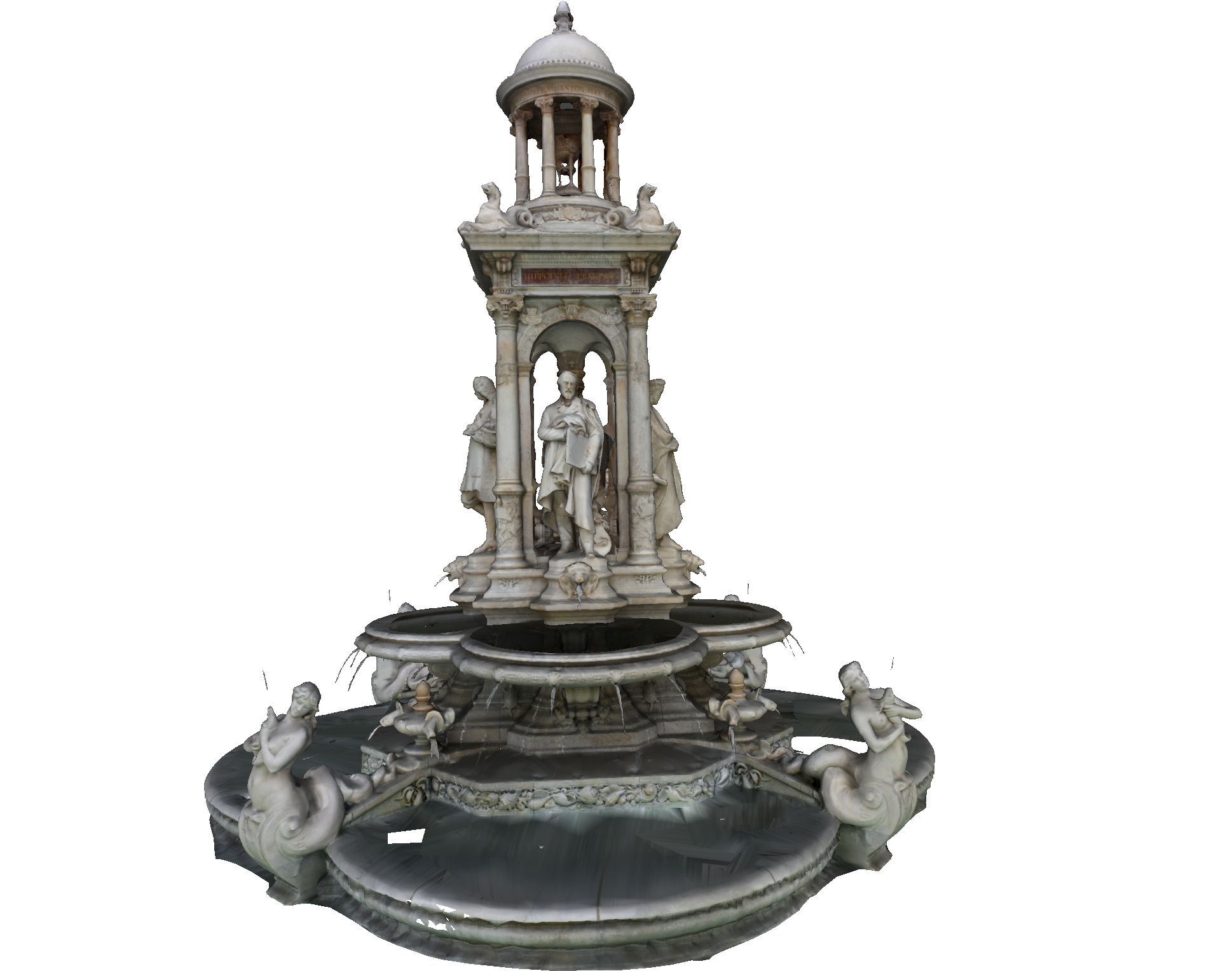}
        \end{subfigure} \\
        \hline 
        \vrulel{\rotatebox[origin=c]{90}{\parbox{2cm}{\centering MegaDepth \\(internet)}}} &
        \begin{subfigure}{0.55\columnwidth}
           	 	\includegraphics[height=68pt]{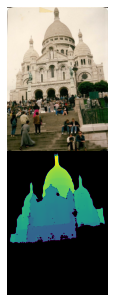}
             	\includegraphics[height=68pt]{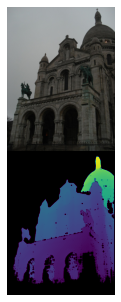}
             	 \includegraphics[height=68pt]{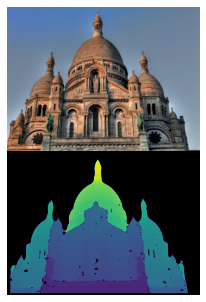}
              	 \includegraphics[height=68pt]{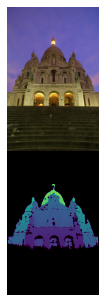}     
       	\end{subfigure} & 
        $\boldsymbol{\cdots}$ & 
       	\vruler{\begin{subfigure}{0.3 \columnwidth}
                \includegraphics[width=\linewidth, trim={200pt, 0, 200pt, 100pt}, clip]{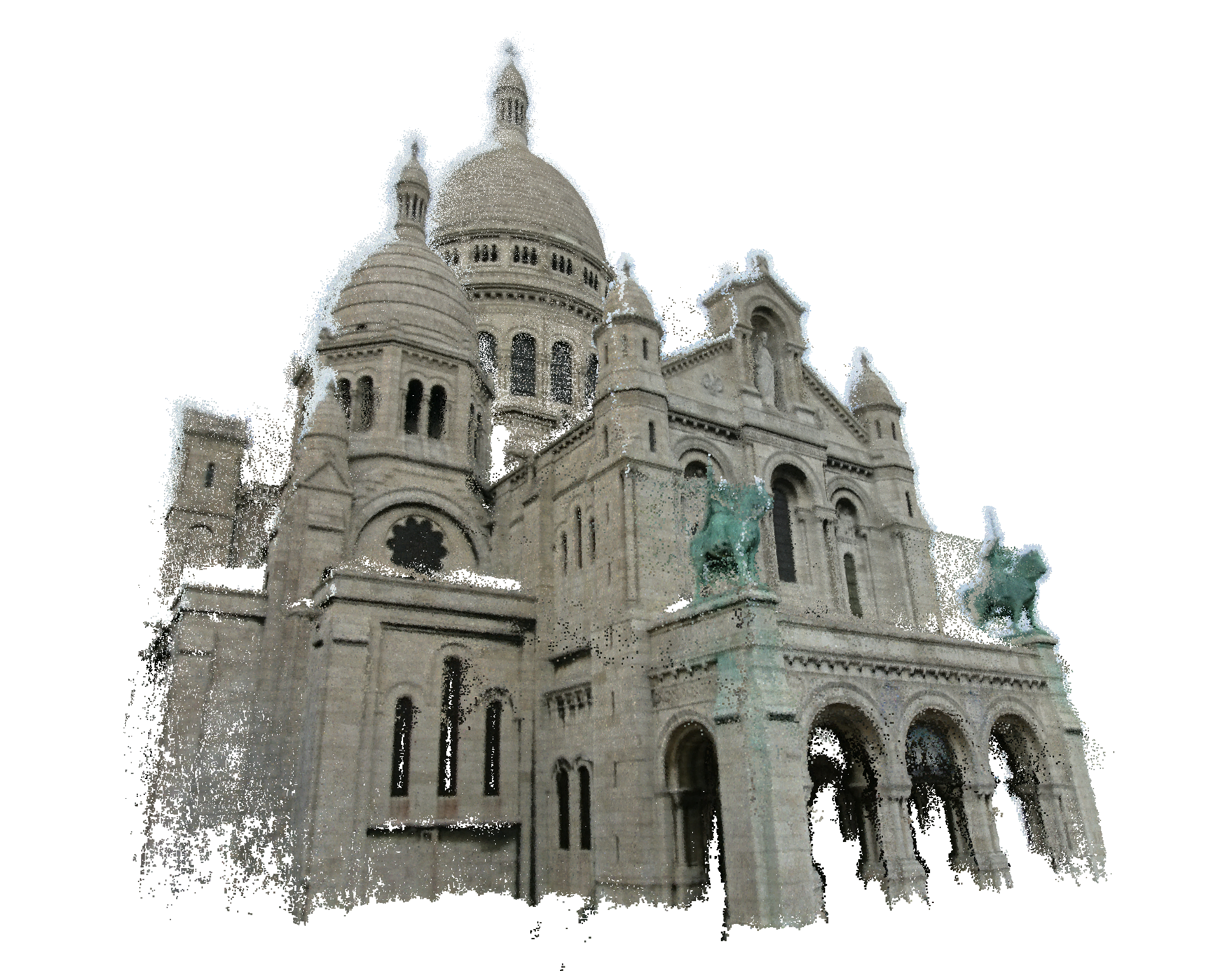}
        \end{subfigure}} \\
        \hline 

    \end{tabular}
    \caption{We apply deep multi-view stereo networks to internet images and study the influence of architecture, supervision and training data over the quality of the reconstructed 3D models.}
    \label{fig:teaser}

\end{figure}

Deep Learning holds promises to improve Multi-View Stereo (MVS), i.e. the reconstruction of a 3D scene from a set of images with known camera parameters. However, deep MVS algorithms have mainly been developed and demonstrated in controlled settings, e.g. using pictures taken with a single camera, at a given time, specifically for 3D reconstruction. In this paper, we point out the challenges in using such approaches to perform 3D reconstructions ``in the wild" 
and propose an evaluation procedure for such a task. We analyze quantitatively and qualitatively the impact of network architecture and training for 3D reconstructions with a varying number of images. We give particular attention to unsupervised methods and propose a new approach which performs on par with state of the art while being much simpler. 



The reference dataset for the development of the vast majority of deep MVS approaches~\cite{yao2018mvsnet,yao2019recurrent,yang2020cost,chen2019point,yu2020fast,luo2019p} has been the {DTU}~\cite{jensen2014large} dataset, consisting of 80 small scenes in sequences of 49 to 64 images. It was acquired in a lab, together with a reference 3D model obtained with a structured light scanner. While the scale of this dataset makes it suitable to train deep networks, it corresponds to a relatively simple reconstruction scenario, where many images taken in similar conditions are available. Recently, deep MVS approaches have started to evaluate their performance on larger outdoor datasets, such as \emph{Tanks and Temples}~\cite{knapitsch2017tanks} (T\&T) and ETH3D~\cite{schoeps2017cvpr}. For example, VisMVSNet~\cite{zhang2020visibility} outperformed all classical methods on the T\&T dataset. Yao \etal~\cite{yao2020blendedmvs} have recently proposed a synthetic dataset, BlendedMVS, specifically designed to train deep MVS approaches to generalize to such outdoor scenarios. We aim to go further and evaluate whether deep MVS networks can generalize to the more challenging setting of 3D reconstruction from Internet image collections. We are particularly interested in the challenging case with few images available for the reconstruction, where classical approaches struggle to reconstruct dense point clouds, and we propose an evaluation protocol for such a setup.

Such a generalization to datasets with different properties, where obtaining ground truth measurements is challenging, has been a major motivation for unsupervised deep MVS approaches~\cite{dai2019mvs2,khot2019learning,huang2020m,xu2021self}. We found however that many of these methods cannot train the network on real data. We thus propose a simpler alternative providing results on par with the best competing methods. It relies on minor modifications of the standard MVSNet~\cite{yao2018mvsnet} architecture and is simply trained by comparing the images warped using the predicted depth.

Our experiments give several interesting insights. 
First, we show that existing unsupervised method are not suited for images in the wild but our proposed method is. Second, we show that depthmap-based MVS methods are state-of-the-art for reconstruction of few internet images. Third, in the wild evaluation provides very different results than evaluation on DTU and T\&T. We therefore advocate for its use when developing new architectures. 
%
Our main contributions are:
\begin{itemize}
    \setlength{\itemsep}{0pt}
    \setlength{\parskip}{0pt}
    \setlength{\parsep}{0pt} 
    \item We introduce a new experimental setup\footnote{Our PyTorch code and data are available at \url{https://github.com/fdarmon/wild_deep_mvs}.} for comparing different algorithms on images in the wild.
    \item We use this protocol to compare methods along three axes: network supervision, network architecture and training dataset.
    \item We show that existing unsupervised methods fail on images in the wild and thus we introduce a new unsupervised method.
\end{itemize}

\section{Related Work}


The goal of MVS is to reconstruct the 3D geometry of a scene from a set of images and the corresponding camera parameters. These parameters are assumed to be known, 
e.g., estimated using structure-from-motion~\cite{schonberger2016structure}.
Classical approaches to MVS can be roughly sorted in three categories~\cite{furukawa2015multi}: direct reconstruction~\cite{furukawa2009accurate} that works on point clouds, volumetric approaches~\cite{seitz1999photorealistic,kutulakos2000theory} that use a space discretization, and depth map based methods~\cite{galliani2015massively,schonberger2016pixelwise} that first compute a depth map for every image then fuse all depth maps into a single reconstruction. The latter approach decomposes MVS in many depth estimation tasks, and is typically adopted by deep MVS methods. 

\vspace{-2mm}
\paragraph{Deep Learning based MVS}

Deep Learning was first applied to patch matching~\cite{zbontar2015computing,zagoruyko2015learning} and disparity estimation~\cite{kendall2017end}. Several approaches tackled the MVS problem with deep learning, using either a volumetric approach~\cite{ji2017surfacenet,atlas} or sparse point cloud densification~\cite{deltas}. MVSNet~\cite{yao2018mvsnet} introduced a network architecture specific to depth map based MVS and end-to-end trainable. Further works improved the precision, speed and memory efficiency of MVSNet. \cite{yao2019recurrent, yan2020dense} propose a recurrent architecture instead of 3D convolutions. MVS-CRF~\cite{xue2019mvscrf} uses a CRF model with learned parameters to infer depth maps. Many approaches introduce a multistep framework where a first depth map is produced with an MVSNet-like architecture, then the results are refined either with optimization~\cite{yao2019recurrent, yu2020fast}, graph neural networks~\cite{chen2019point,chen2020visibility} or multiscale neural networks~\cite{gu2020cascade,yang2020cost,zhang2020visibility, yi2020pyramid}. Another line of work aims at improving the per-view aggregation by predicting aggregation weights~\cite{luo2019p, zhang2020visibility}. All these methods were evaluated in controlled settings and it is unclear how they would generalize to internet images. 

Unsupervised training is a promising approach for such a generalization. \cite{khot2019learning,dai2019mvs2,huang2020m, xu2021self} propose unsupervised approaches using several photometric consistency and regularization losses. However, to the best of our knowledge, these methods were never evaluated on Internet-based image datasets. Besides, their loss formulation is composed of many terms and hyperparameters, making them difficult to reproduce and likely to fail on new datasets. In this paper we propose a simpler unsupervised training approach, which we show to perform on par with these more complex ones on standard data, and generalizes to Internet images. 

\vspace{-2mm}
\paragraph{Datasets and evaluation}

There exist many datasets for 3D reconstruction. Most use active sensing~\cite{geiger2013vision,knapitsch2017tanks, strecha2008benchmarking}. Synthetic ones~\cite{sintel} provide an alternative way to acquire large-scale data but they cannot be used to predict reliable performance on real images.
Specific datasets were introduced for MVS, such as ETH3D~\cite{schoeps2017cvpr}, Tanks \& Temples (T\&T)~\cite{knapitsch2017tanks} or DTU~\cite{jensen2014large} datasets.
Most deep MVS methods train on the DTU dataset~\cite{jensen2014large} and evaluate generalization on T\&T~\cite{knapitsch2017tanks}. DTU provides large-scale data but in a laboratory setting. The limited generalization capability of networks trained on DTU was highlighted in~\cite{yao2020blendedmvs}. In this paper, a network trained on the synthetic dataset {BlendedMVS} was shown to perform better on T\&T than its counterpart trained on MegaDepth~\cite{li2018megadepth}. \cite{yao2020blendedmvs} claims that this is due to the lack of accuracy on MegaDepth depth maps. We also compare training deep MVS approaches on DTU, BlendedMVS and MegaDepth but for MVS reconstruction from Internet images, and additionally compare performances obtained in supervised and unsupervised settings. 

\section{MVS Networks in the Wild}

In this section, we present standard deep depthmap-based MVS architectures and training which we compare, and propose some modifications, including our new unsupervised training, to better handle images in the wild.  

\subsection{Network Architectures}

\begin{figure}
    \centering
    \includegraphics[width=0.8\columnwidth]{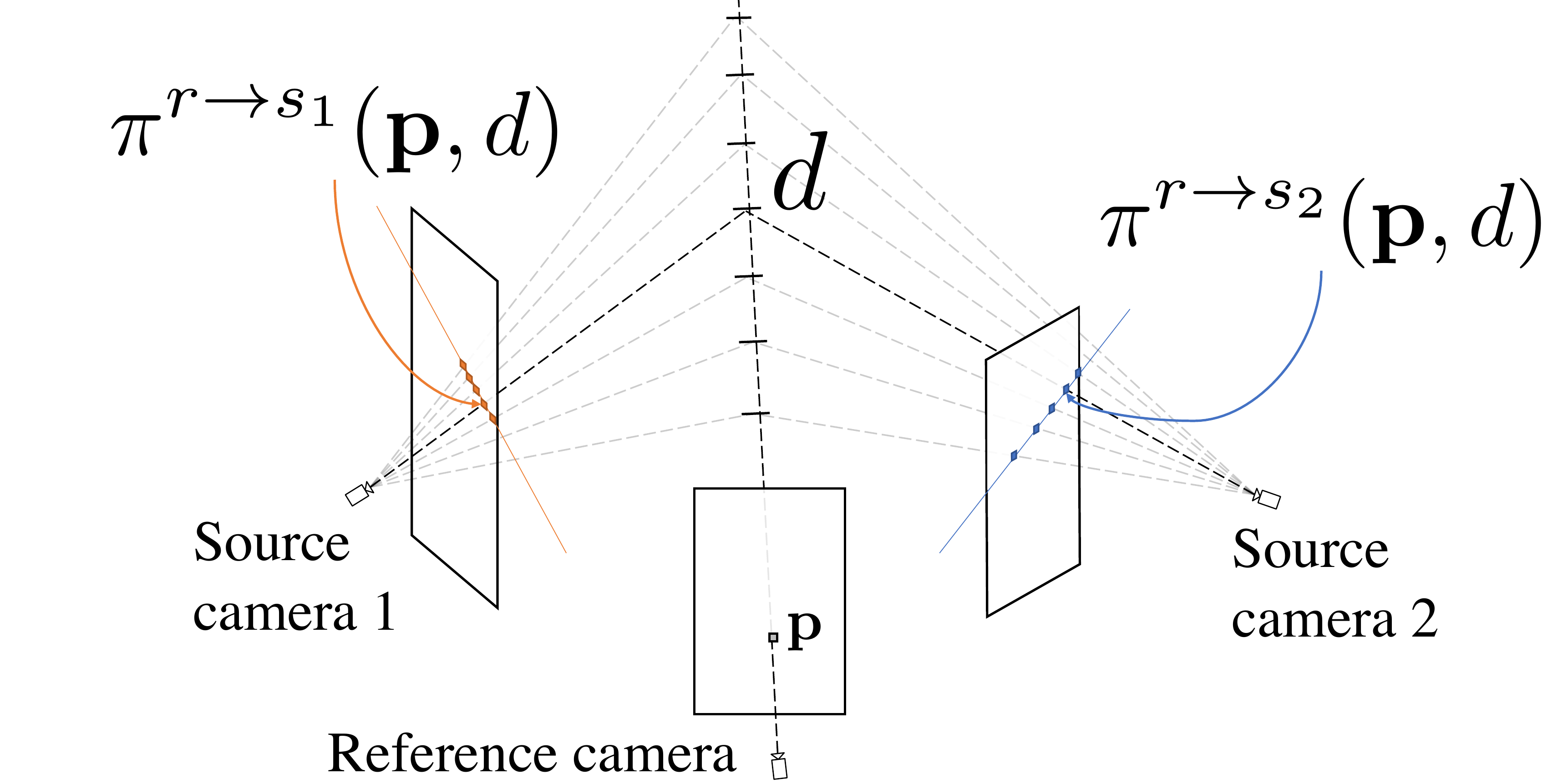}
    \caption{Notations: A pixel $\*{p}$ in reference frame is projected on source frames at a position depending on its depth.}
    \label{fig:cost_volume}
\end{figure}



\paragraph{Overview and notations}

Starting from a set of calibrated images, deep MVS methods typically try to estimate a depth map for each image. 
The depth maps are then fused into a 3D point cloud using an independent method. 
The MVS network will take as input a \emph{reference image} $\*I^r$ and a set of \emph{source images}  $\*I^{s_1}\dots \*I^{s_{N}}$, and try to predict the depth~$\*D^r$ of the reference image. 

Multi-view stereo architectures build on the idea that depth can be used to establish correspondences between images. More formally
, as visualized in Figure~\ref{fig:cost_volume}, we write $\*{\pi}^{\rs{k}}(\*p, d)$ the 2D position of the projection in the camera associated to $\*I^{s_k}$ of the 3D point at depth $d$ on the ray associated to the pixel $\*p\in\mathbb{R}^2$ in $\*I^r$. If $d$ is the ground truth depth of the pixel $\*p$ then the appearance of the reference image should be consistent with all the appearances of the source images $\*I^{s_k}$ at pixels $\*{\pi}^{\rs{k}}(\*p, d)$. 


Similar to the classic plane sweep stereo methods, the best performing deep MVS methods explicitly sample candidate depths and build 
 a cost volume that computes for every pixel a score based on the correspondences given in every view by the candidate depths. This score can be seen as the cost of assigning a given depth to a pixel. The depth map can be extracted from the cost volume by looking at the minimum in the depth axis. 
 MVSNet~\cite{yao2018mvsnet} was the first deep architecture implementing this idea.
\vspace{-2mm}
\paragraph{MVSNet}

MVSNet starts by extracting a feature representation of each input image using a shared convolutional neural network (CNN). We write $\*F^r$ the feature map of the reference image and $\*F^{s_1} \dots \*F^{s_{N}}$ the feature maps of the source images. The network also takes as input the depth range $[d_{\min},d_{\max}]$ to consider. It is usually provided in standard datasets, and we explain in section~\ref{sec:dmin} how we defined them in the context of reconstruction in the wild. The information from the feature maps are then aggregated into a 3D cost volume $C$ for depth $d$ regularly sampled from $d_{\min}$ to $d_{\max}$.
\begin{equation}
    \mathbf{C}[\*p, d] = \phi \left(\*F^r[\*p]; \{\*F^{s_k}[\*{\pi}^{\rs{k}}(\*p, d)] ,k = 1 \dots N\} \right),
\end{equation}
where $\*F[\cdot]$ denotes the interpolation of the feature map $\*F$ and $\phi$ is an aggregation function. In MVSNet the channel-wise variance is used for aggregation:
\begin{equation}
    \phi(\*f_r; \{\*f_{s_1}, \dots \*f_{s_{N}}\}) = \text{var}(\*f_r, \*f_{s_1}, \dots \*f_{s_{N}}).
    \label{eq:phimvsnet}
\end{equation}
The cost volume only encodes local information. A further step of 3D convolutions is then performed to propagate this local information. The depth map is finally extracted from the refined cost volume using a differentiable argmin operation.

\vspace{-2mm}
\paragraph{Softmin-based aggregation}

The variance aggregation of MVSNet can be problematic when applied to images in the wild. Indeed, we can expect large viewpoint variations and many occlusions. Both would lead to large appearance differences in the pixels corresponding to the actual depth, and such outlier images would have a strong influence on the variance. Instead, we propose an alternative aggregation: 
\begin{multline}
    \phi(\*f_r; \{\*f_{s_1}, \dots \*f_{s_{N}}\}) = \\
    \frac{\sum_{k=1}^{N}\exp(-\lambda\|\*f_r - \*f_{s_k}\|^2) (\*f_r - \*f_{s_k})^2}{\sum_{k=1}^{N}\exp(-\lambda \|\*f_r - \*f_{s_k}\|^2)},
    \label{eq:phiours}
\end{multline}
where $\lambda$ is a learnable parameter. This score has two main differences with the variance aggregation. First, it breaks the symmetry between the reference and source images, each source feature is compared to the reference instead of the mean feature. Second, in our aggregation, the contribution of each source feature is weighted using a softmin weight. The intuition is that the features most similar to the ones of the reference image should be given higher weight compared to features that are completely different, which might be outliers. We will show in our experiments that such an aggregation is particularly beneficial in the presence of a high number of potential source images. 

\vspace{-2mm}
\paragraph{Multi-scale architectures.}

A known limitation of MVSNet is its large memory consumption. The consequence is that the resolution, both for image and depth is limited by the available memory. Several methods have been proposed to increase the resolution without increasing the memory requirements. Among them are the multiscale approaches that first scan coarsely the whole depth range using low resolution feature maps then refine the depth at higher resolutions. We used two successful multi-scale approaches: CVP-MVSNet~\cite{yang2020cost} and Vis-MVSNet~\cite{zhang2020visibility}. CVP-MVSNet starts from a downsampled version of the images to predict a coarse depth map using the same architecture as MVSNet. Then, at each scale, this depth map is upsampled and refined using the same network until full scale prediction is reached. This architecture is very light and can work with a variable number of scales. Vis-MVSNet uses a fixed number of scales with a different network for each scale. Moreover, contrary to CVP-MVSNet, the lowest scale features are extracted from the full scale image using strided convolutions. Finally, Vis-MVSNet leverages a specific multi-view aggregation. It first predicts a confidence for each source image using a Bayesian framework~\cite{kendall2017what}, then uses the confidence to aggregate the pairwise cost volumes associated with every source image. While this aggregation is complex, we believe it is one of the key ingredients that allows Vis-MVSNet to generalize well to Internet images.  

\subsection{Training}

\subsubsection{Supervised Training.} We first consider the case where a partial ground truth depth map $D_{GT}$ is available, together with a binary mask $M_{GT}$ encoding pixels for which the ground truth is valid. 
Supervised training is typically performed using the $\ell 1$ loss between the predicted depth map $D$ and the ground truth depth map $D_{GT}$, weighted by the mask $M_{GT}$ and normalized by the scene's depth range, 
%
\begin{equation}
    l(D,D_{GT},M_{GT}) = \frac{\|M_{GT}(D- D_{GT})\|}{(d_{\max} -d_{\min}) \|M_{GT}\|}.
\end{equation}
%

\subsubsection{Unsupervised Training}


Motivated by the difficulty to obtain ground truth depth maps, unsupervised approaches attempt to rely solely on camera calibration. Approaches proposed in the literature typically combine a photometric consistency term with several additional losses. Instead, we propose to rely exclusively on the structural similarity between the reference image and the warped source images. 

Similarly to MVS$^2$~\cite{dai2019mvs2}, we adopt a multi-reference setting where each image in a batch is successively used as a reference image. Let us consider a batch of images $\mathcal{I}=\{\*I^1,\ ..., \*I^{N+1}\}$ and the associated depth maps $\mathcal{D}=\{D^1,\ ..., D^{N+1}\}$ predicted by our MVS network. We write $\*W^{\ba} ( D^a)$ the image resulting from warping $\*I^b$ on $\*I^a$ using the estimated depth map $D^a$:

\begin{equation}
    \*W^{\ba} (D^a)[\*p] = \*I^b[\pi^{\ab}(\*p, D^a[\*p])].
\end{equation}
Note that this warped image is not defined everywhere, since the 3D point defined from the pixel $\*p$ and the depth  $D^a[\*p]$ in $\*I^a$ might project outside of $\*I^b$. Even when the warped image is defined, it might not be photoconsistent in case of occlusion. We thus follow MVS$^2$~\cite{dai2019mvs2} and define an occlusion aware mask in $\*I^a$, $M^{\ba}(D^a, D^b)$, based on the consistency between the estimated depths $D^a$ and $D^b$ in $I^a$ and $I^b$ (see~\cite{dai2019mvs2} for details).


We then define the loss for the batch as:
\vspace{-2mm}
\begin{multline}
    l(\mathcal{I},\mathcal{D}) = \\ \sum_{r=1}^{N} \frac{\sum_{s\neq r} \left\|M^{\sr{}}(D^r,D^s) \text{SSIM}\left(\*I_r, \*W^{\sr{}}(D^r)\right)\right\|}{\sum_{s\neq r}\left\|M^{\sr{}}(D^r,D^s) \right\|}
    \label{eq:occ_loss}
\end{multline}
%
%
 where SSIM is the structural similarity~\cite{ssim}, which enables to compare images acquired in very different conditions.
To avoid the need for further regularization outlined by other unsupervised approaches, we use the same approach as~\cite{shen2020ransac}: the depth maps, which are predicted at the same resolution as the CNN feature maps, are upsampled to the resolution of the original images before computing the loss. 

Note that this loss is much simpler than in MVS$^2$~\cite{dai2019mvs2}, a combination of $\ell 1$ color loss, $\ell 1$ of the image gradients, SSIM and census transform. However, we show that it leads to results on par with state of the art and can be successfully applied to train a network from Internet image collections.


\section{Benchmark of Deep MVS Networks in the Wild}
To analyze the performance of different networks, we define training and evaluation dataset, evaluation metrics, as well as  reference choices for several steps of the reconstruction pipeline, including reference depth to consider and a strategy to merge estimated depths into a full 3D model.

\subsection{Data}

\paragraph{Training datasets}
\label{sec:dmin}

We experiment training on the DTU~\cite{jensen2014large}, BlendedMVS~\cite{yao2020blendedmvs} and MegaDepth~\cite{li2018megadepth} datasets. Training an MVS network requires images with their associated calibration and depth maps. It also requires to select the set of source images for each reference image and the depth range for building the cost volume. 
DTU is a dataset captured in laboratory conditions with structured light scanner ground truth. It was preprocessed, organized, and annotated by Yao~\etal~\cite{yao2018mvsnet}, who generated depth maps by meshing and rendering the ground truth point cloud.
BlendedMVS was introduced in order to train deep models for outdoor reconstruction. It is composed of renderings of 3D scenes blended with real images to increase realism. Depth range and list of source views are given in the dataset. These datasets have reliable ground truth but there is a large domain gap between their images and Internet images. Thus, we also experiment training on MegaDepth, which is composed of many Internet images with pseudo ground truth depth and sparse 3D models obtained with COLMAP~\cite{schonberger2016pixelwise,schonberger2016pixelwise}. We generate a training dataset from MegaDepth, excluding 
the training scenes of the Image Matching Benchmark~\cite{Jin_2020} which we will use for evaluation. 
We selected sets of training images by looking at the sparse reconstruction. We randomly sample a reference image and two source images uniformly among all the images that have more than 100 reconstructed 3D points, with a triangulation angle above 5 degree, in common with the reference.
Once these sets of three images are sampled, we select the 3D points observed by at least three images and use their minimum and maximum depth as $d_{\min}$ and $d_{\max}$. 
\vspace{-2mm}
\paragraph{Depth map evaluation}
To evaluate depth map prediction, we used the synthetic BlendedMVS scenes, as well as the images from YFCC-100M~\cite{thomee2016yfcc100m, heinly2015reconstructing} used in the image matching benchmark of~\cite{Jin_2020}. Indeed, \cite{Jin_2020} provides 14~sets of few thousands of images with filtered ground truth depth maps obtained with COLMAP~\cite{schonberger2016structure,schonberger2016pixelwise}.

\vspace{-2mm}
\paragraph{3D reconstruction evaluation}
Because our ultimate goal is not to obtain depth maps but complete 3D reconstructions, we also select data to evaluate it. We used the standard DTU test set, but also constructed an evaluation setting for images in the wild. 
We start from the verified depth maps of the Image Matching Workshop~\cite{Jin_2020}. We fuse them into a point cloud with COLMAP fusion using very conservative parameters: reprojection error below half a pixel and depth error below $1\%$. The points that satisfy such conditions in $20$ views are used as reference 3D models. We manually check that the reconstructions were of high enough quality. The scenes \textit{Westminster Abbey} and \textit{Hagia Sophia interior} were removed since their reconstruction was not satisfactory. We then randomly selected sets of 5, 10, 20, and 50 images as test image sets and compare the reconstructions using only the images in these sets to the reference reconstruction obtained from the reference depth maps of a very large number of images. 
While the reference reconstructions are of course biased toward COLMAP reconstructions, we argue that the quality and extension of the reconstructions obtained from the small image sets is much lower than the reference one, and thus evaluating them with respect to the reference makes sense. Moreover, we never observed any case where part of the scene was reconstructed using deep MVS approaches on the test image sets and not in the reference reconstruction. 

\subsection{Metrics}

\paragraph{Depth estimation}

We follow the same evaluation protocol as BlendedMVS~\cite{yao2020blendedmvs}. The predicted and ground truth depth maps are first normalized by $(d_{\max} - d_{\min})/128$ which makes the evaluation comparable for images seeing different depth ranges. We use the following three metrics: end point error (EPE), the mean absolute error between the inferred and the ground truth depth maps; $\mathbf{e_1}$, the proportion in \% of pixels with an error larger than 1 in the scaled depth maps; $\mathbf{e_3}$, the proportion in \% of pixels with an error larger than 3. 

\vspace{-2mm}
\paragraph{3D reconstruction}
On DTU~\cite{jensen2014large} we follow the standard evaluation protocol and use as metrics:  Accuracy, the mean distance from the reconstruction to ground truth, completeness from ground truth to reconstruction, and the overall metric defined as the average of the previous two. 

For our evaluation in the wild, we use the same metrics as T\&T~\cite{knapitsch2017tanks} : precision, recall and F-score at a given threshold. The reference reconstruction is known only up to a scale factor therefore the choice of such threshold is not straightforward. Let $D_k$ be the ground truth depth map of image $I_k$, we define the threshold as:
\begin{equation}
    t = \underset{k}{\text{median}}\left(\underset{\|\*p - \*p'\| = 2}{\text{median}}\left \|D_k[\*p] \*p - D_k[\*p'] \*p'\right\|\right).
\end{equation}

\subsection{Common Reconstruction Framework}

To fairly compare different deep MVS approaches for 3D reconstruction, we need to fuse the predicted depth maps in a consistent way. 
Several fusion methods exist, each with variations in the point filtering and the parameters used. For DTU reconstructions, we use fusibile~\cite{galliani2015massively} with the same parameters for all compared methods. Since fusibile is only implemented for fixed resolution images, we use a different method for Internet images. We use the fusion procedure of COLMAP~\cite{schonberger2016pixelwise} with a classic pre-filtering step similar to the one used in MVSNet: we only keep depths consistent in three views. The consistency criterion is based on three thresholds: the reprojected depth must be closer than $1\%$ to original depth, the reprojection distance must be less than 1 pixel, and the triangulation angle must be above 1\degree. 

\section{Experiments}

We first provide an ablation study of our unsupervised approach and compare it with state of the art unsupervised methods (Section~\ref{sec:expe_unsup}). Then, we compare the different network architecture and supervisions for depth map estimation (Section~\ref{sec:expe_depth}) and full 3D reconstruction (Section~\ref{sec:expe_3D}).  

\subsection{Unsupervised Approach Analysis}
\label{sec:expe_unsup}
We validate and ablate our unsupervised approach on both DTU 3D reconstruction and YFCC depth map estimation in Table~\ref{tab:ablation}. We test the importance of upsampling the depth map as a way to regularize the network, compare variance and softmin aggregation functions as well as results with and without occlusion aware masking. We also compare our approach with existing work.\footnote{We use open source implementations of M3VSNet and JDACS provided by the authors.}
The main observation is that upsampling is the key factor to obtain meaningful results. Softmin aggregation is consistently better than variance for both datasets and evaluations. Occlusion masking also consistently improves the results, by a stronger margin on YFCC where occlusions are expected to be more common. Using upsampling and occlusion masking, our simple training loss is on par with state of the art unsupervised methods on DTU but is is much better for YFCC trainings where competing method have poor results. 

\begin{table}[t] 
    \footnotesize
    \tabcolsep=0.15cm
    \centering
    \caption{Ablation study of unsupervised learning. We compare networks trained with and without upsampling (Up.) of the predicted depth map, with and without occlusion masking (Occ.) and using as aggregation function ($\phi$) either Variance aggregation (V, eq.~\ref{eq:phimvsnet}) or our Softmin aggregation (S, eq.~\ref{eq:phiours}).}
    \label{tab:ablation}
    \begin{tabular}{c c c|c c c|c c c}
        \multirow{2}{*}{$\phi$} & \multirow{2}{*}{Up.} & \multirow{2}{*}{Occ.} & \multicolumn{3}{c|}{DTU reconstruction} & \multicolumn{3}{c}{YFCC depth maps}\\
        & & & Prec.  & Comp. & Over. & EPE & $e_1$ & $e_3$ \\
        \hline
        V & & & 1.000 & 0.803 & 0.901 & 34.74 & 91.88 & 80.99 \\
        V & \checkmark & & 0.614 & 0.580 & 0.597 & 21.68 & 67.48 & 48.44 \\
        S & \checkmark & & \bf 0.607 & 0.560 & 0.584 & 21.88 & 66.39 & 46.61 \\
        V & \checkmark & \checkmark & 0.610 & 0.545 & 0.578 & 18.75 & 63.07 & 43.17 \\
        S & \checkmark & \checkmark & 0.608 & \bf 0.528 & \bf 0.568 & \bf 18.22 & \bf 61.97 & \bf 41.34 \\
        \hline
        \multicolumn{3}{c|}{Unsup. MVSNet\cite{khot2019learning}} & 0.881 & 1.073 & 0.977 \\
        \multicolumn{3}{c|}{MVS$^2$~\cite{dai2019mvs2}} & 0.760 &  \bf 0.515 & 0.637 \\
        \multicolumn{3}{c|}{M$^3$VSNet~\cite{huang2020m}} & 0.636 & 0.531 & 0.583 & 40.57 & 82.45 & 69.91\\
        \multicolumn{3}{c|}{JDACS~\cite{xu2021self}} & \underline{\bf 0.571} & \underline{\bf 0.515} & \underline{\bf 0.543} & 34.37 & 92.36 & 80.45

    \end{tabular}
    \vspace{-10pt}
\end{table}

\subsection{Depth Map Prediction}
\label{sec:expe_depth}
In Table~\ref{tab:depth map_comp} we compare different approaches 
for depth prediction on the YFCC data and BlendedMVS validation set. We study different architectures, the standard MVSNet architecture with either the standard variance aggregation or our proposed softmin aggregation and two state-of-the-art multiscale architecture CVP-MVSNet~\cite{yang2020cost} and Vis-MVSNet~\cite{zhang2020visibility}.\footnote{For both CVP-MVSNet and Vis-MVSNet, we adapted the public implementation provided by the authors} We compare results for networks trained on DTU, BlendedMVS (B) and MegaDepth (MD) with either the supervised $\ell 1$ loss or our occlusion masked unsupervised loss (except for BlendedMVS which we only use in the supervised setting since it is a synthetic dataset designed specifically for this purpose). 

\vspace{-2mm}
\paragraph{MVSNet architecture} 
As expected, the best performing networks on a given dataset are the ones trained with a supervised loss on the same dataset. Confirming our analysis in the unsupervised setup, we can also see that MVSNet modified with our softmin-based aggregation systematically outperforms the variance-based aggregation. 
The results for unsupervised settings are more surprising. First, networks trained on DTU in the unsupervised setting generalize better both to blended-MVS and YFCC, hinting that unsupervised approaches might lead to better generalization even without changing the training data.
Also, the unsupervised network trained on MegaDepth performs better on YFCC than the supervised networks trained on BlendedMVS, outlining again the potential of unsupervised approaches. 

\begin{table}[t]
    \footnotesize
    \tabcolsep=0.19cm
    \centering
    \caption{Direct depth map evaluation: Comparison of architectures MVSNet~\cite{yao2018mvsnet}, our MVSNet with softmin aggregation (MVSNet-S), CVP-MVSNet~\cite{yang2020cost} and Vis-MVSNet~\cite{zhang2020visibility}, trained on BlendedMVS (B), DTU or MegaDepth (MD), with or without depth supervision (Sup.).}
    \label{tab:depth map_comp}
    \begin{tabularx}{\linewidth}{c c c|c c c|c c c}
        \multirow{2}{*}{\rotatebox[origin=c]{90}{Archi}} & \multirow{2}{*}{\rotatebox{90}{Train}} & \multirow{2}{*}{\rotatebox[origin=c]{90}{Sup.}} & \multicolumn{3}{c|}{BlendedMVS} & \multicolumn{3}{c}{YFCC }\\
        & & & EPE & $e_1$ & $e_3$ & EPE & $e_1$ & $e_3$ \\
        \hline
        \multirow{5}{*}{\rotatebox[origin=c]{90}{MVSNet}} & B & \checkmark & \bf 1.49 & \bf 21.98 & \bf  8.32 & 21.56 & 67.93 & 49.75 \\
        & DTU & \checkmark & 4.90 & 39.25 & 22.01 & 32.41 & 79.74 & 65.42 \\
        & DTU & & 3.94 & 31.68 & 16.61 & 24.61 & 70.77 & 53.44 \\
        & MD & \checkmark & 2.29 & 30.08 & 12.37 & \bf 17.61 & \bf 62.54 & 43.09 \\
        & MD & & 3.24 & 32.94 & 15.25 & 18.75 & 63.07 & \bf 42.17 \\
        \hline
        \multirow{5}{*}{\rotatebox[origin=c]{90}{MVSNet-S}} & B & \checkmark & \bf 1.35 & \bf 25.91 & \bf 8.55 & 20.98 & 69.57 & 49.86 \\
        & DTU & \checkmark & 4.18 & 42.07 & 19.87 & 27.18 & 78.25 & 60.53 \\
        & DTU & & 4.05 & 32.50 & 16.01 & 21.07 & 68.64 & 50.35 \\
        & MD & \checkmark & 1.91 & 32.91 & 11.93 & \bf 15.57 & \bf 60.44 & \bf 38.18 \\
        & MD & & 2.88 & 33.44 & 14.14 & 18.22 & 61.97 & 41.34 \\
        \hline
        \multirow{5}{*}{\rotatebox[origin=c]{90}{CVP-MVSNet}} & B & \checkmark & \bf 1.90 & \bf 19.73 & \bf 10.24 & 40.07 & 85.88 & 76.25 \\
        & DTU & \checkmark & 10.99 & 46.79 & 35.59 & 73.69 & 95.29 & 90.69 \\
        & DTU & & 5.25 & 29.33 & 19.77 & 45.36 & 89.81 & 82.17 \\
        & MD & \checkmark & 3.07 & 24.33 & 14.40 & 34.39 & \bf 78.49 & \bf 66.57 \\
        & MD & & 3.39 & 21.67 & 12.94 & \bf 32.74 & 78.92 & 66.99 \\
        \hline
        \multirow{5}{*}{\rotatebox[origin=c]{90}{Vis-MVSNet}} & B & \checkmark & \bf 1.47 & \bf 18.47 & \bf 7.59 & 19.60 & 64.98 & 46.38 \\
        & DTU & \checkmark & 3.70 & 30.37 & 18.16 & 27.46 & 72.89 & 58.37 \\
        & DTU & & 7.22 & 37.03 & 20.75 & 38.32 & 73.24 & 56.17 \\
        & MD & \checkmark & 2.05 & 22.21 & 10.14 & \bf 16.01 & \bf 56.71 & \bf 38.20 \\
        & MD & & 3.88 & 31.64 & 17.50 & 19.21 & 66.27 & 47.34 \\
    \end{tabularx}
    \vspace{-10pt}
\end{table}

\vspace{-2mm}
\paragraph{Multi-scale architectures}
CVP-MVSNet, which is our best performing architecture on DTU (see next section) and achieves the second best results in terms of $e_1$ error on BlendedMVS, did not achieve satisfactory results on YFCC images even when supervised on MegaDepth. This may be because at the lowest resolution the architecture takes as input a low resolution downsampled version of the image, on which correspondences might be too hard to infer for images taken in very different conditions. 

On the contrary, Vis-MVSNet supervised on MegaDepth data is the best performing approach in terms of $e_1$ error. However, its performance is worse in the unsupervised setting compared to MVSNet. This is likely explained by the complexity of the loss function of Vis-MVSNet. In the original paper, the loss combines pairwise estimated depth maps and final depth maps. The pairwise loss is computed in a Bayesian setting~\cite{kendall2017what} and our direct adaptation of this loss by replacing $\ell 1$ with photometric loss might be too naive.

\begin{figure}[t]
    \centering
    \includegraphics[width=0.85\linewidth]{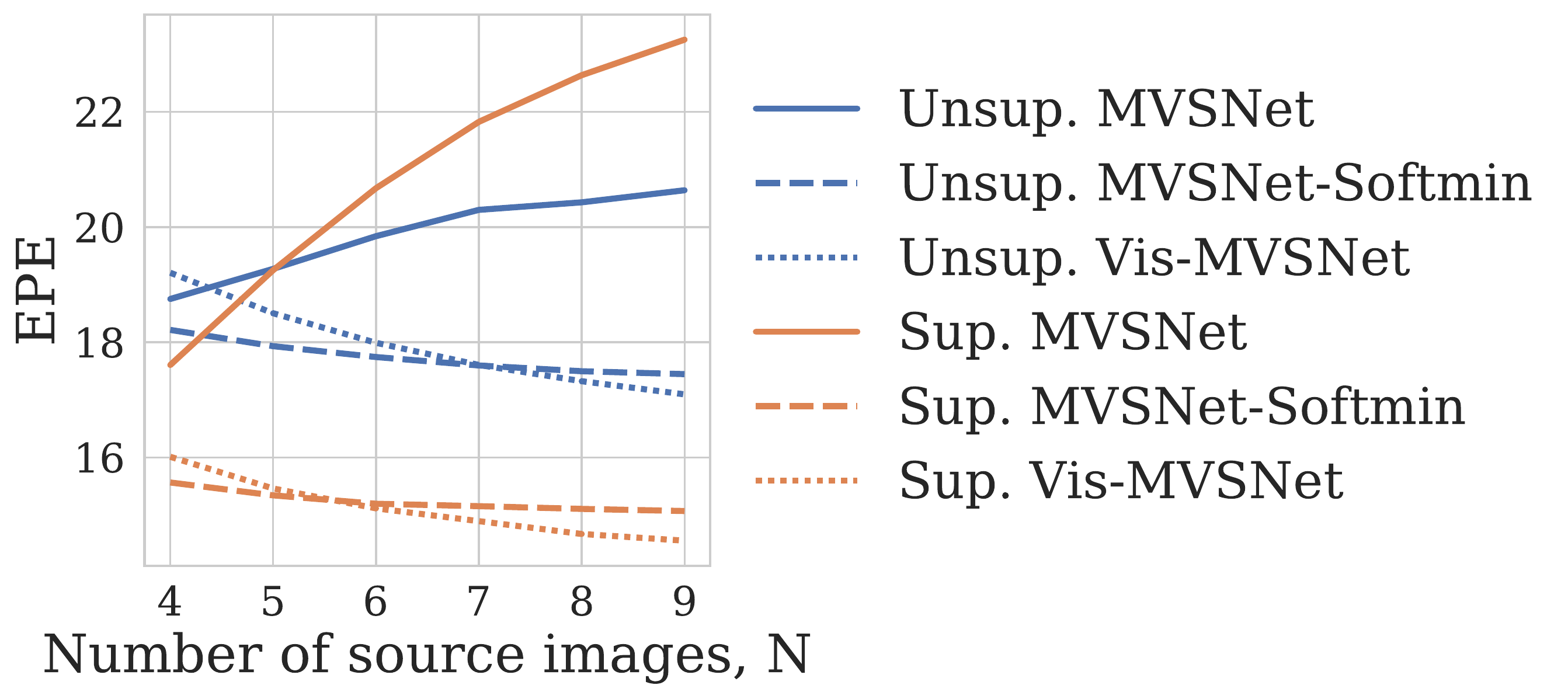}
    \caption{EPE on YFCC depth map estimation as a function of the number of source images $N$ for networks trained on MegaDepth, supervised (Sup.) or unsupervised (Unsup.). Contrary to the other architectures MVSNet with variance aggregation does not benefit from using more images.}
    \label{fig:md_fun_N}
    \vspace{-10pt}
\end{figure}

\vspace{-2mm}
\paragraph{Number of source views}
Until now, all experiments were performed using one reference view and four source views as input to the network at test time. However, in real scenarios, one might have access to more views, and it is thus important for the network to benefit from an increased number of views at inference time. We thus investigate the use of more source images at test time, still training with two source images only, and report the results in Figure~\ref{fig:md_fun_N}. The results for MVSNet architecture with variance-based aggregation get worse as the number of source views is increased. This may be because the relevant information in the cost volume is more likely to be masked by noise when the number of source images increases. On the contrary, the performances of our softmin-based aggregation or Vis-MVSNet aggregation improve when using a higher number of source views.

\subsection{3D Reconstruction}
\label{sec:expe_3D}
We now evaluate the quality of the full 3D reconstruction obtained with the different architectures. The predicted depth maps are combined with a common fusion algorithm and the evaluation is performed on the fused point cloud. We first report the results on DTU in Table~\ref{tab:3D_comp_classical}. Interestingly, our softmin-based adaptation degrades performance when training on DTU in a supervised setting, but not in an unsupervised setting. Also note that the performance on DTU, and in particular the completeness of the reconstruction, is largely improved when the networks are trained on MegaDepth data. The best performance is even obtained in an unsupervised setting with CVP-MVSNet. However, this network performs poorly on YFCC data.

\begin{figure*}[ht]
    \renewcommand{\arraystretch}{.5}
    \begin{tabular}{c|c|ccccc}
        \multirow{3}{*}{\rotatebox[origin=c]{90}{\hspace{-5cm}MVSNet~\cite{yao2018mvsnet}}} &
        \rotatebox[origin=c]{90}{Blended Sup.} &
        \raisebox{-0.5\height}{\includegraphics[width=0.12\linewidth, trim={100pt 0 60pt 0}, clip]{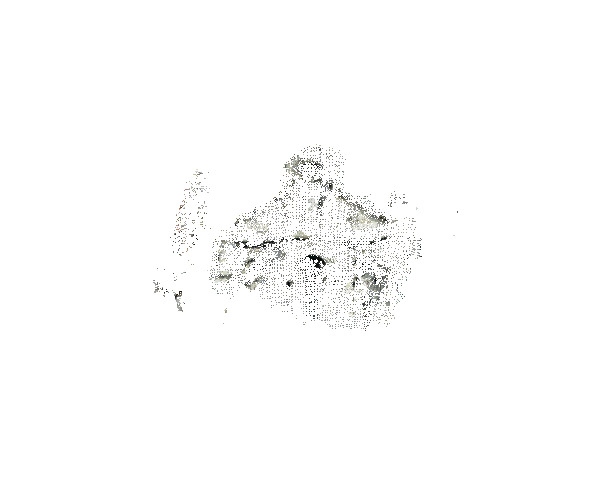}} &
        \raisebox{-0.5\height}{\includegraphics[width=0.16\linewidth]{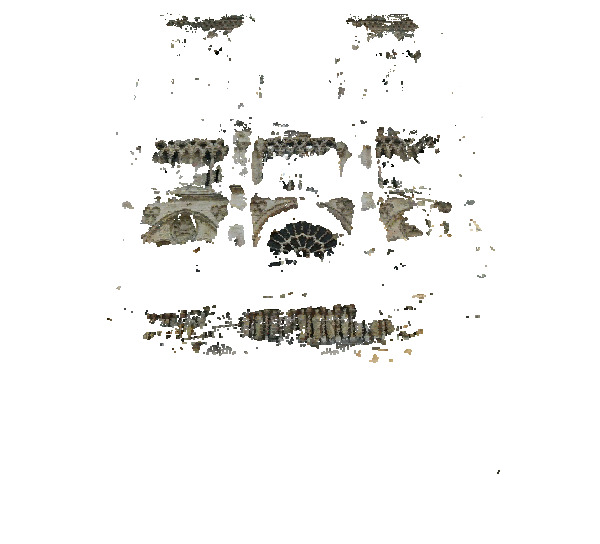}} &
        \raisebox{-0.5\height}{\includegraphics[width=0.16\linewidth]{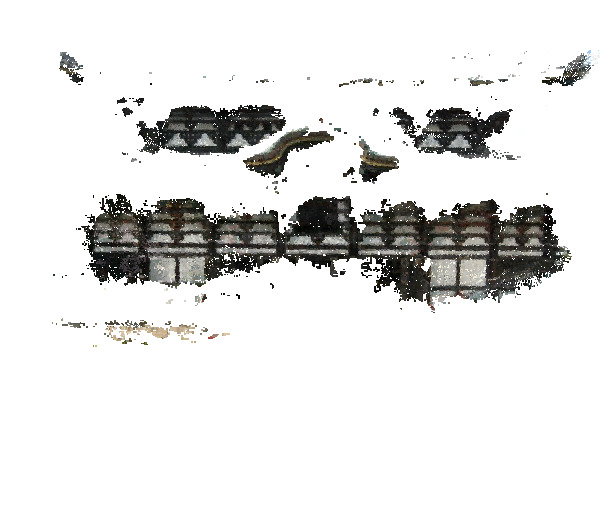}} &
        \raisebox{-0.5\height}{\includegraphics[width=0.16\linewidth]{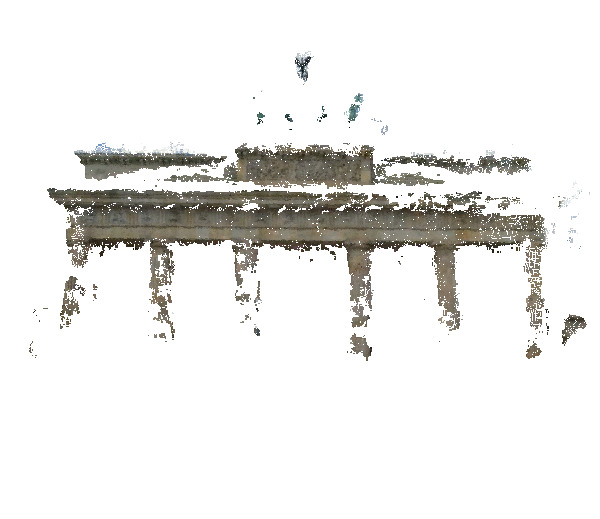}} &
        \raisebox{-0.5\height}{\includegraphics[width=0.16\linewidth]{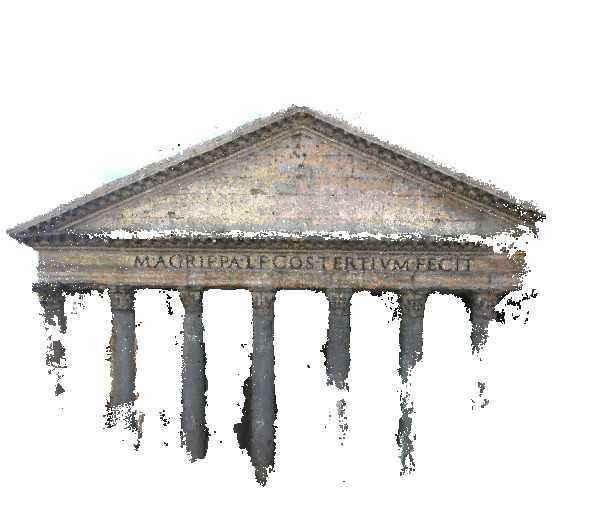}}
        \\
        \cline{2-2}
        &
        \rotatebox[origin=c]{90}{MD Sup.} &
        \raisebox{-0.5\height}{\includegraphics[width=0.12\linewidth, trim={100pt 0 60pt 0}, clip]{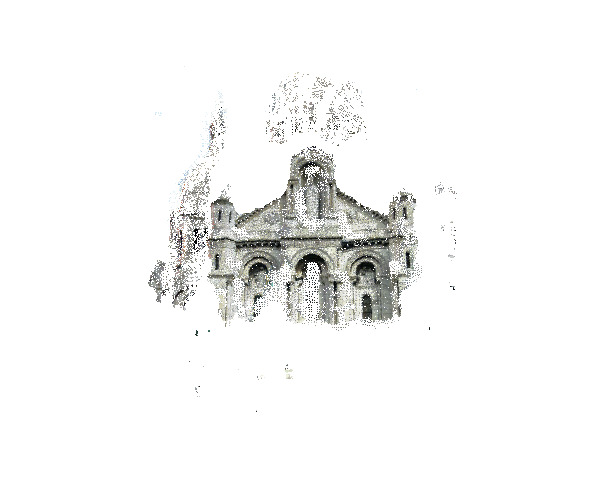}} &
        \raisebox{-0.5\height}{\includegraphics[width=0.16\linewidth]{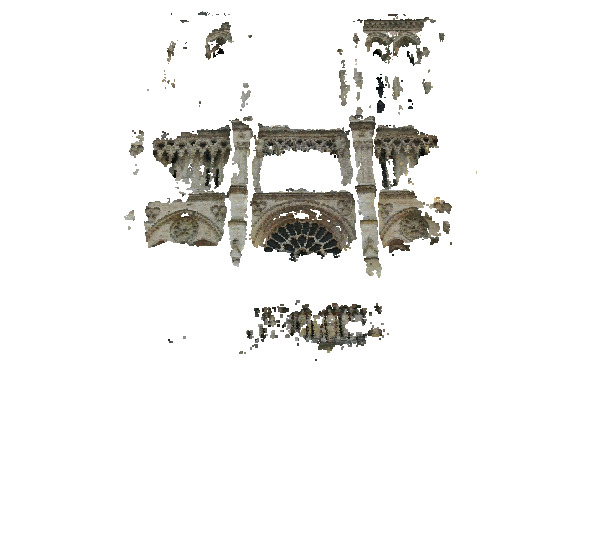}} &
        \raisebox{-0.5\height}{\includegraphics[width=0.16\linewidth]{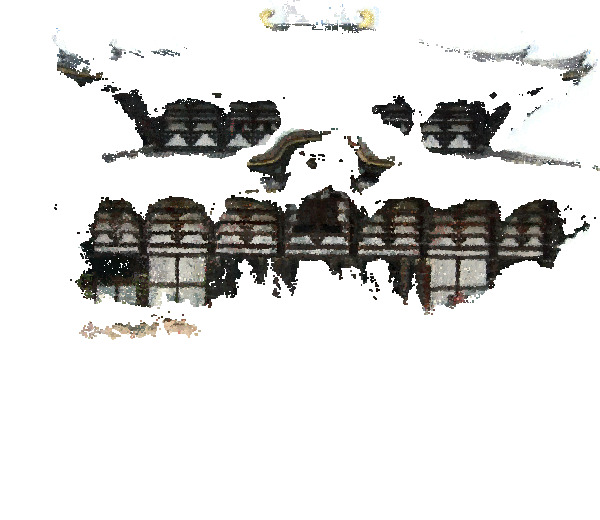}} &
        \raisebox{-0.5\height}{\includegraphics[width=0.16\linewidth]{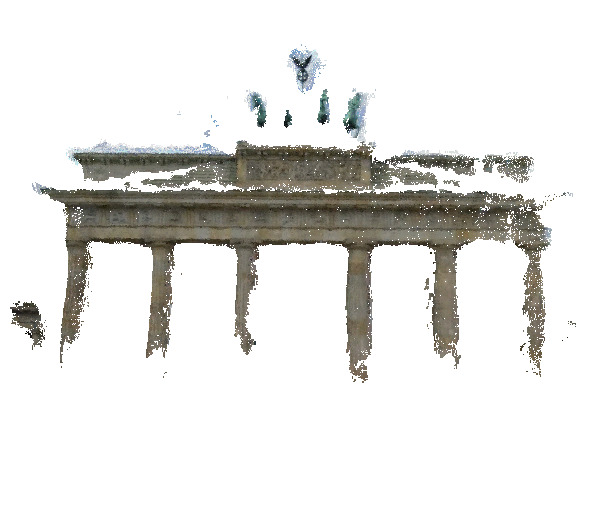}} &
        \raisebox{-0.5\height}{\includegraphics[width=0.16\linewidth]{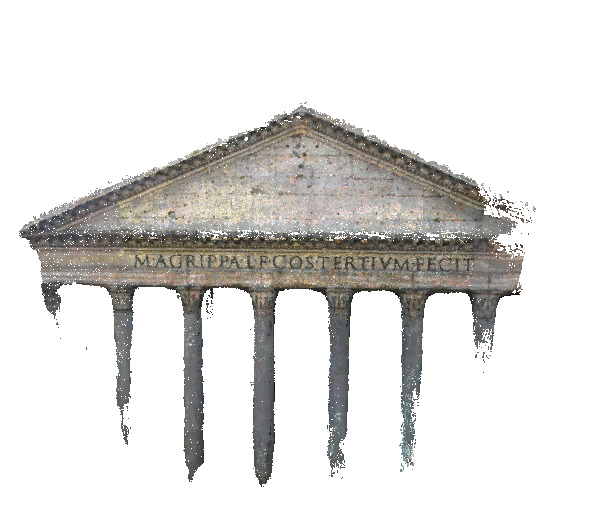}}
        \\
        \cline{2-2}
        &
        \rotatebox[origin=c]{90}{MD Unsup.} &
        \raisebox{-0.5\height}{\includegraphics[width=0.12\linewidth, trim={100pt 0 60pt 0}, clip]{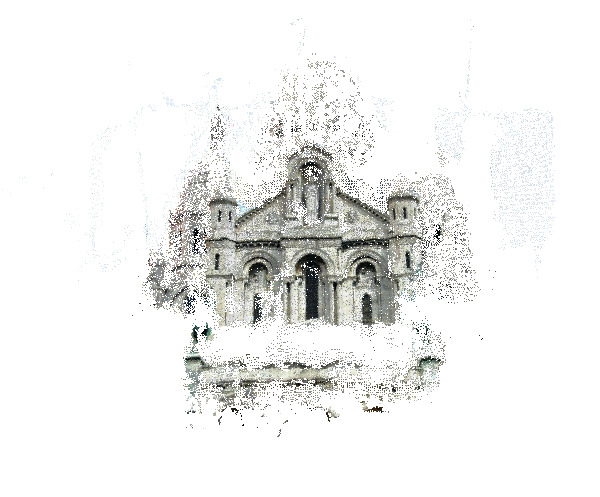}} &
        \raisebox{-0.5\height}{\includegraphics[width=0.16\linewidth]{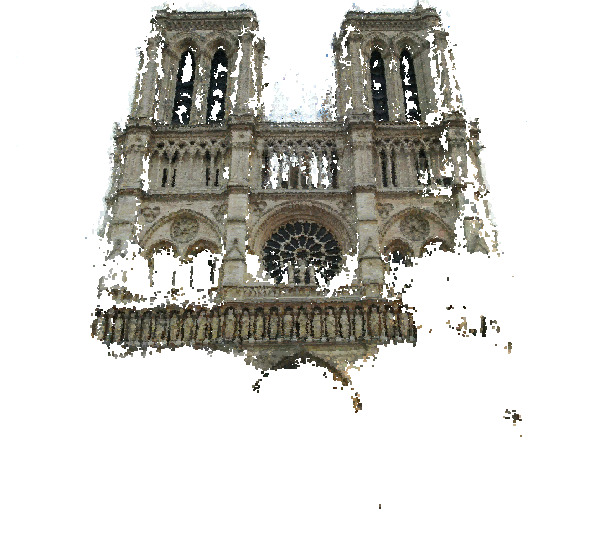}} &
        \raisebox{-0.5\height}{\includegraphics[width=0.16\linewidth]{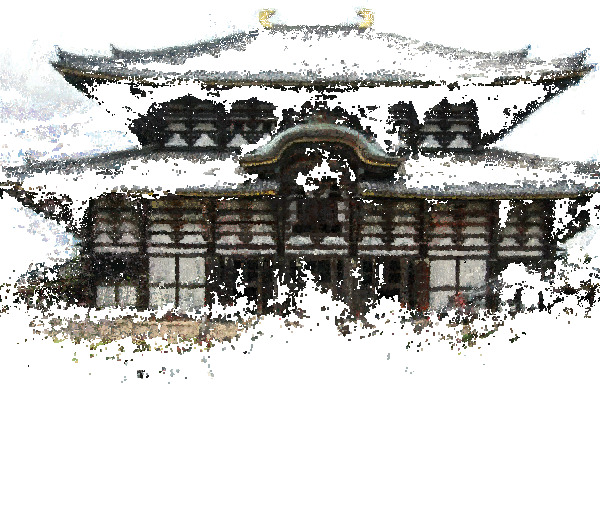}} &
        \raisebox{-0.5\height}{\includegraphics[width=0.16\linewidth]{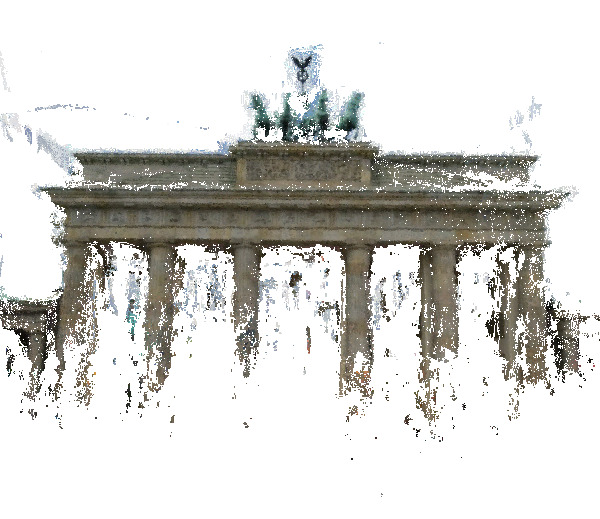}} &
        \raisebox{-0.5\height}{\includegraphics[width=0.16\linewidth]{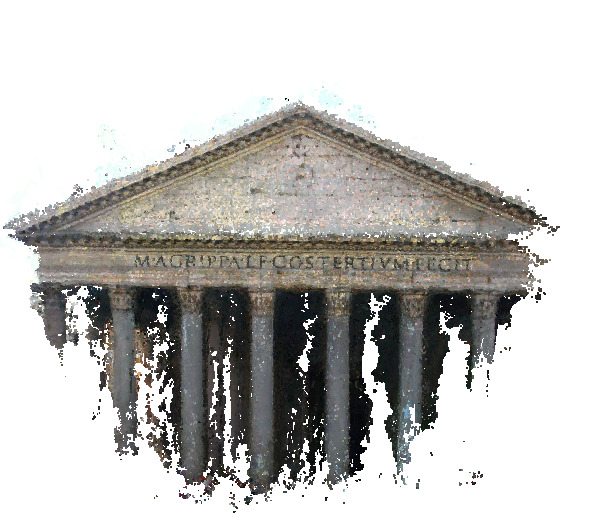}}
        \\
        \cline{1-2}
        \rotatebox[origin=c]{90}{\small Vis-MVSNet~\cite{zhang2020visibility}} &
        \rotatebox[origin=c]{90}{ MD Sup.} &
        \raisebox{-0.5\height}{\includegraphics[width=0.12\linewidth, trim={100pt 0 60pt 0}, clip]{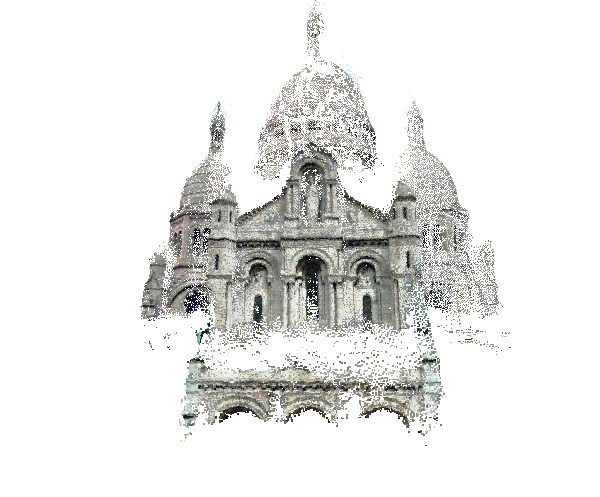}} &
        \raisebox{-0.5\height}{\includegraphics[width=0.16\linewidth]{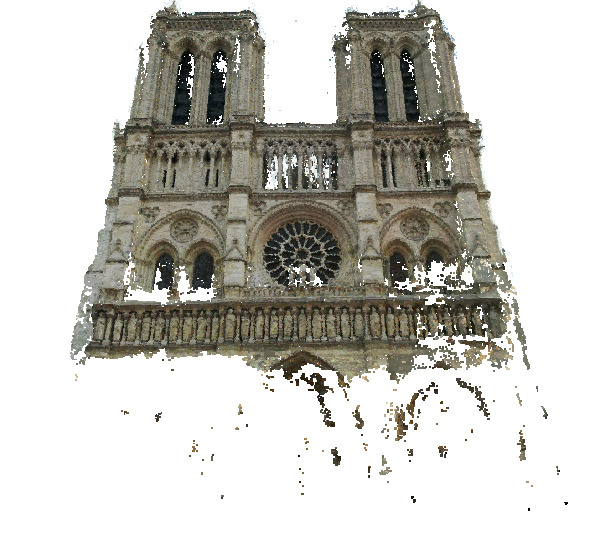}} &
        \raisebox{-0.5\height}{\includegraphics[width=0.16\linewidth]{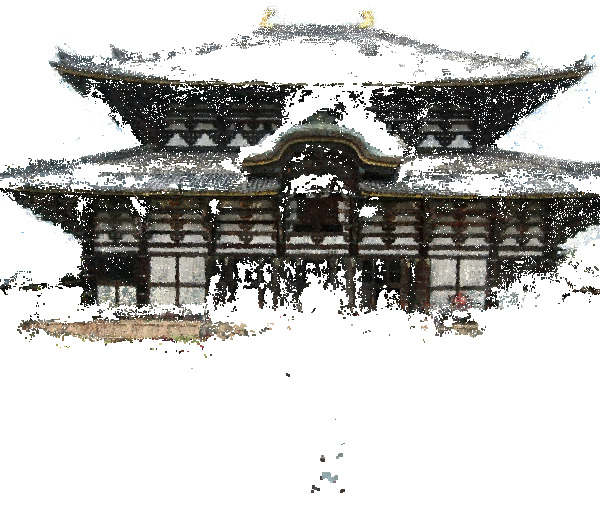}} &
        \raisebox{-0.5\height}{\includegraphics[width=0.16\linewidth]{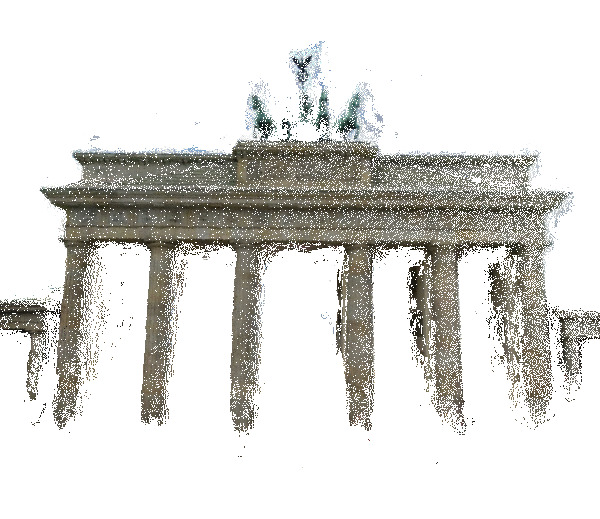}} &
        \raisebox{-0.5\height}{\includegraphics[width=0.16\linewidth]{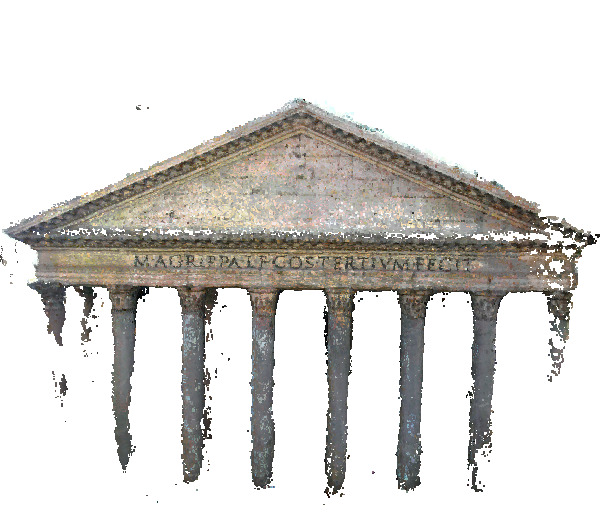}}
        \\
        \cline{1-2}
        \multicolumn{2}{c|}{\rotatebox[origin=c]{90}{COLMAP~\cite{schonberger2016pixelwise}}} &
        \raisebox{-0.5\height}{\includegraphics[width=0.12\linewidth, trim={100pt 0 60pt 0}, clip]{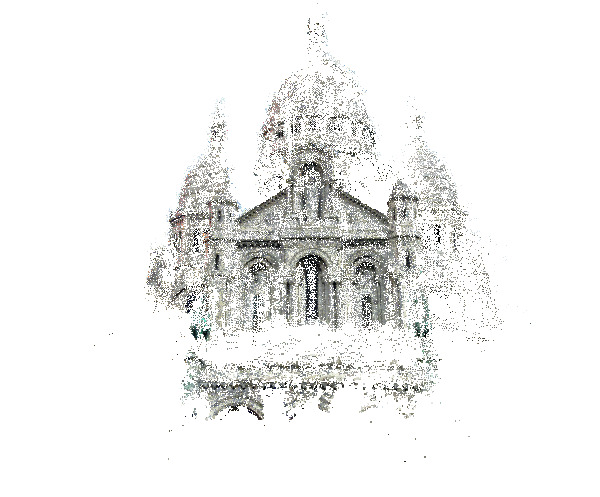}} &
        \raisebox{-0.5\height}{\includegraphics[width=0.16\linewidth]{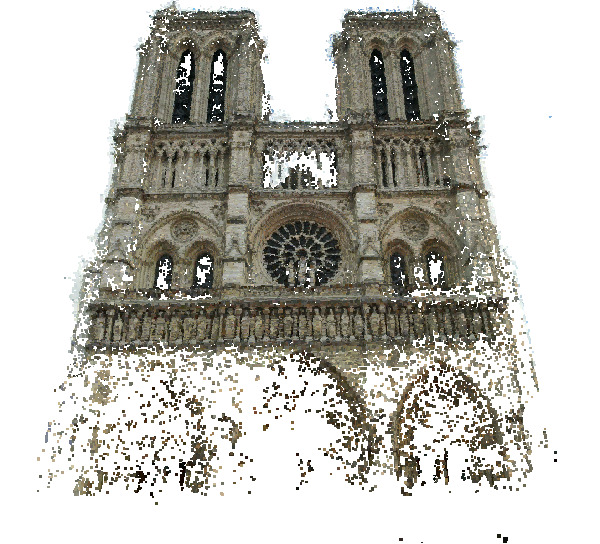}} &
        \raisebox{-0.5\height}{\includegraphics[width=0.16\linewidth]{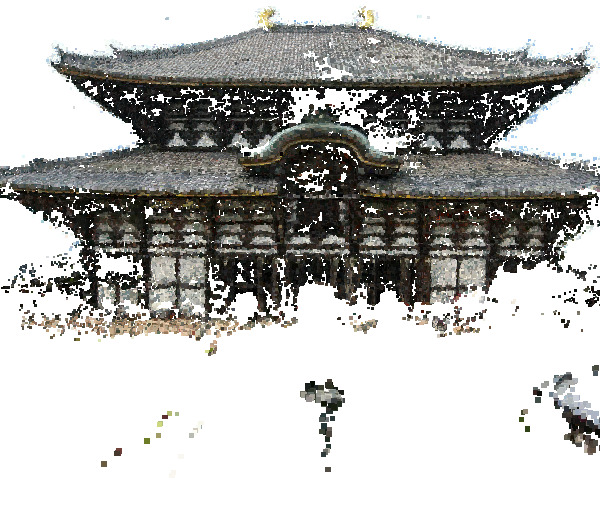}} &
        \raisebox{-0.5\height}{\includegraphics[width=0.16\linewidth]{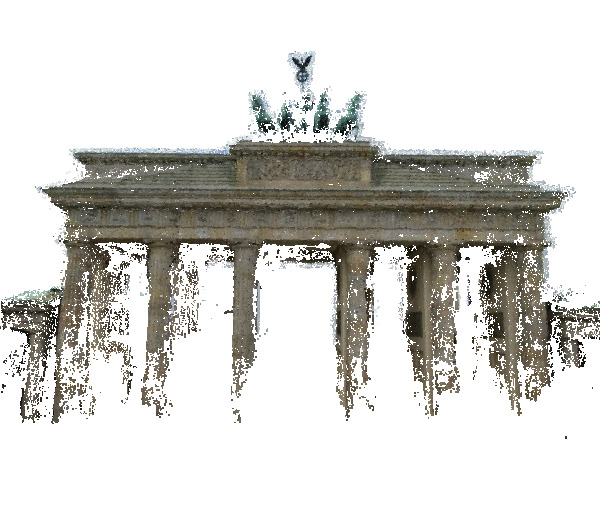}} &
        \raisebox{-0.5\height}{\includegraphics[width=0.16\linewidth]{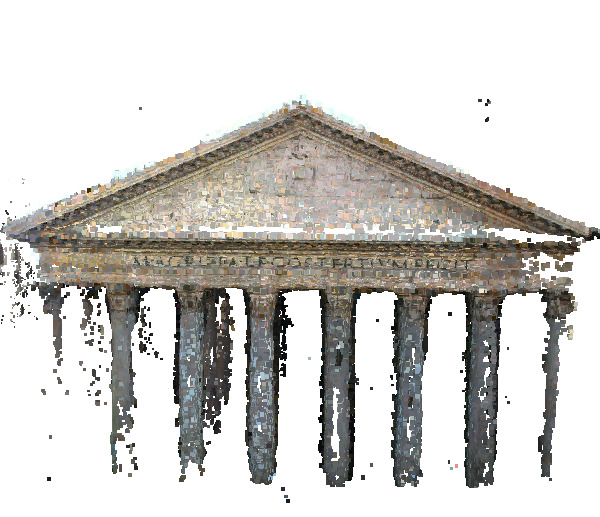}}
        \\
        \hline
        \multicolumn{2}{c|}{\rotatebox[origin=c]{90}{Ground Truth}} &
        \raisebox{-0.5\height}{\includegraphics[width=0.12\linewidth, trim={100pt 0 60pt 0}, clip]{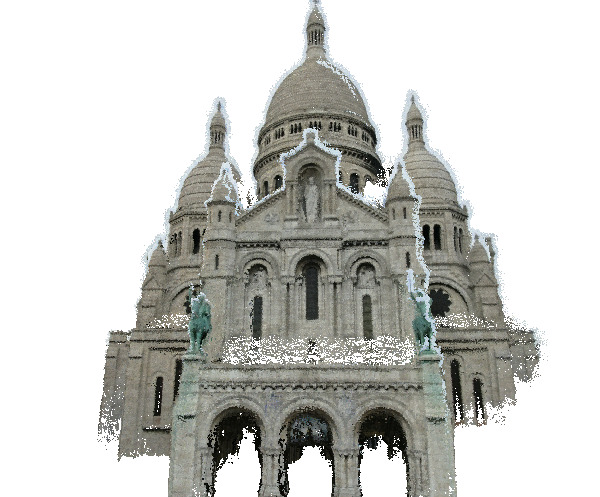}} &
        \raisebox{-0.5\height}{\includegraphics[width=0.16\linewidth]{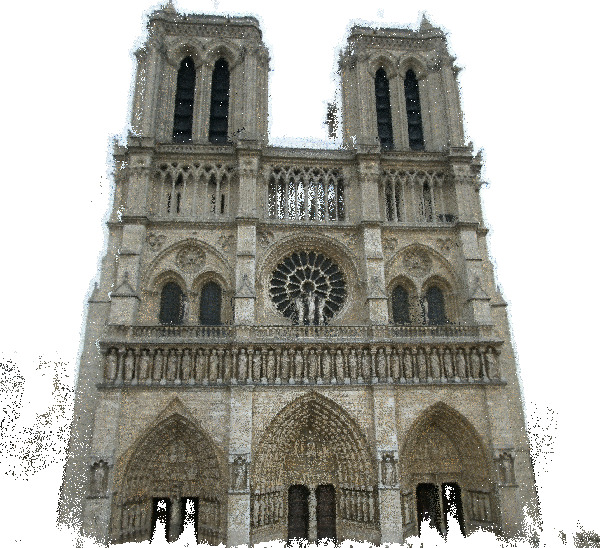}} &
        \raisebox{-0.5\height}{\includegraphics[width=0.16\linewidth]{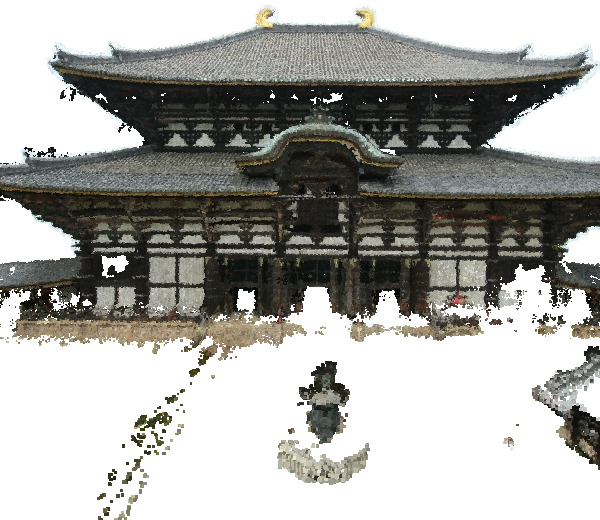}} &
        \raisebox{-0.5\height}{\includegraphics[width=0.16\linewidth]{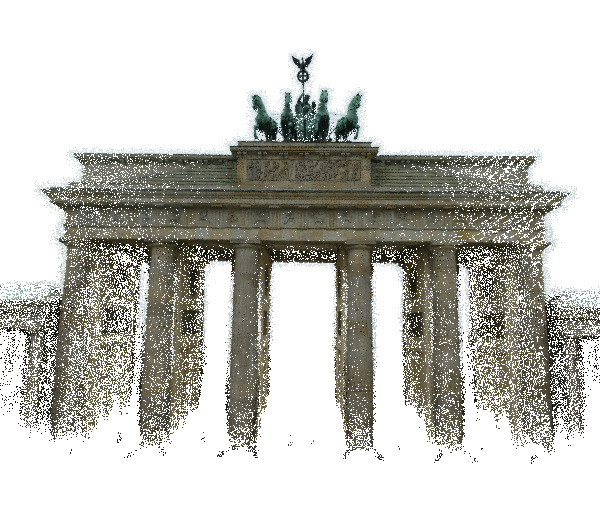}} &
        \raisebox{-0.5\height}{\includegraphics[width=0.16\linewidth]{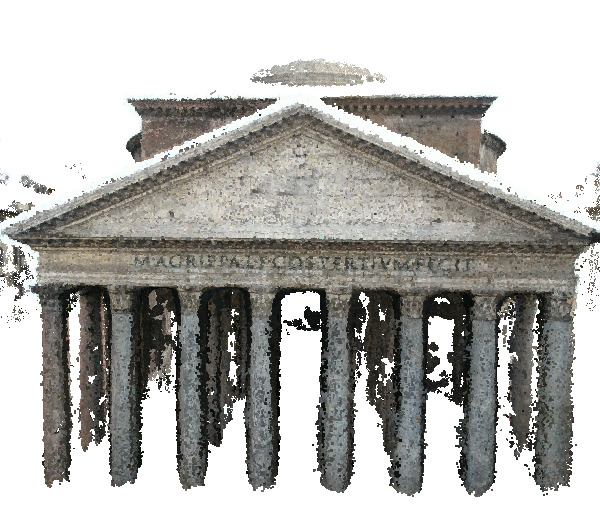}}
        \\
    \end{tabular}
    \caption{Comparison of reconstructions from ten images using different architectures trained on BlendedMVS (Blended) or MegaDepth (MD), supervised (Sup.) or unsupervised (Unsup.). }
   \label{fig:quali_examples}
   \vspace{-10pt}
\end{figure*}

Finally, we evaluate the quality of the 3D reconstructions obtained from small sets of YFCC images. For fair comparison with COLMAP~\cite{schonberger2016pixelwise}, the depth maps are evaluated with up to 19 source images for these experiments, except when using the original MVSNet architecture with variance aggregation, which does not benefit from using more images and which we continue to use with four source images. Since depth estimation on YFCC images completely failed with CVP-MVSNet as well as with networks trained on DTU, we did not perform reconstructions in these cases. 
Quantitative results are reported in Table~\ref{tab:3D_comp_itw} and qualitative results are shown in Figure~\ref{fig:quali_examples}. 

Most of the trends observed for depth map prediction can also be observed in this experiment. In particular, networks trained on MegaDepth perform better than those trained on BlendedMVS, our softmin-based aggregation improves results for the MVSNet architecture, in particular when many views are available and in the unsupervised setting, and Vis-MVSNet outperforms MVSNet in the supervised setting. 

A new important observation is that the unsupervised MVSNet performs better than the supervised version, especially for the recall metric. This can also be seen in Figure~\ref{fig:quali_examples} where the reconstructions appear more complete in the unsupervised setting. We argue that this is a strong result that advocates for the development of unsupervised approaches suitable for the more advanced architectures. Interestingly, this result was not noticeable in our depth map evaluation. 

Another significant observation is that Vis-MVSNet trained with supervision on MegaDepth quantitatively outperforms COLMAP when using very small image sets (5 or 10 images) with both higher precision and higher recall. This is also a very encouraging result, showing that Deep MVS approaches can provide competitive results in challenging conditions. However, one should note that this quantitative result is not obvious when looking at the qualitative results where either COLMAP or Vis-MVSNet achieves better looking results, depending on the scene.

\begin{table}[t]
    \footnotesize
    \centering
    \caption{3D reconstruction evaluation on DTU. Comparison of trainings on BlendedMVS, MegaDepth and DTU for various architectures with or without supervision (`Sup.' column). }
    \label{tab:3D_comp_classical}
    \begin{tabularx}{\columnwidth}{X X c|c c c}
        \multirow{2}{*}{Archi.} & \multirow{2}{*}{\shortstack[c]{Training \\ data}} & \multirow{2}{*}{Sup.} & \multicolumn{3}{c}{DTU reconstructions}\\
        & & & Acc. & Comp. & Overall\\
        \hline
        \multirow{5}{4em}{MVSNet~\cite{yao2018mvsnet}} & Blended & \checkmark & 0.487 & 0.496 & 0.491 \\
        & DTU & \checkmark & \bf 0.453 & \bf 0.488 & \bf 0.470  \\
        & DTU & & 0.610 & 0.545 & 0.578 \\
        & MD & \checkmark & 0.486 & 0.547 &	0.517\\
        & MD & & 0.689 & 0.645 & 0.670\\
        \hline
        \multirow{5}{5em}{\shortstack[c]{MVSNet~\cite{yao2018mvsnet} \\ softmin}} & Blended & \checkmark & 0.631 & 0.738 & 0.684\\
        & DTU & \checkmark & \bf 0.598 & 0.531 & \bf 0.564 \\
        & DTU & & 0.609 & \bf 0.528 & 0.568\\
        & MD & \checkmark & 0.625 & 0.548 & 0.586\\
        & MD & & 0.690 & 0.614 & 0.652\\
        \hline
        \multirow{5}{4em}{CVP-MVSNet~\cite{yang2020cost}} & Blended & \checkmark & 0.364 & 0.434 & 0.399\\
        & DTU & \checkmark & 0.396 & 0.534 & 0.465\\
        & DTU & & \bf 0.340 & 0.586 & 0.463\\
        & MD & \checkmark & 0.360 & 0.376 & 0.368\\
        & MD & & 0.364 & \bf 0.370 & \bf 0.367\\
        \hline
        \multirow{5}{4em}{Vis-MVSNet~\cite{zhang2020visibility}} & Blended & \checkmark & 0.504 & 0.411 & 0.458 \\
        & DTU & \checkmark & \bf 0.434 & 0.478 & 0.456\\
        & DTU & & 0.456 & 0.596 & 0.526\\
        & MD & \checkmark & 0.438 & \bf 0.345 & \bf 0.392 \\
        & MD & & 0.451 & 0.873 & 0.662 \\
    \end{tabularx}
    \vspace{-10pt}
\end{table}

\begin{table*}
\footnotesize
\tabcolsep=0.15cm
\centering
\caption{3D reconstruction evaluation on our `in the wild' benchmark. We compare several architectures trained with and without supervision (`Sup.' column) on BlendedMVS (Blended) and MegaDepth (MD). We report the best between using four source views (*) and all source views (up to 19).}
    \label{tab:3D_comp_itw}
    \begin{tabular}{c c c|c c c|c c c|c c c|c c c}
        \multirow{2}{*}{Archi.} & \multirow{2}{*}{\shortstack[c]{Training \\ data}} & \multirow{2}{*}{Sup.} & \multicolumn{3}{c|}{5 images} & \multicolumn{3}{c|}{10 images} & \multicolumn{3}{c}{20 images} & \multicolumn{3}{c}{50 images}\\
        & & & Prec. & Rec. & F-score & Prec. & Rec. & F-score & Prec. & Rec. & F-score & Prec. & Rec. & F-score \\
        \hline
        \multirow{3}{5em}{MVSNet~\cite{yao2018mvsnet} } & Blended & \checkmark & \bf 84.68 & 11.51 & 0.1916 &  \bf 91.37* & 23.27* & 0.3583* & \bf 91.46* & 36.46* & 0.5173*	 &  \bf 93.40*	& 51.50* & 0.6608*\\
        & MD & \checkmark & 84.31 & 14.84 & 0.2387 & 91.05* & 27.62* & 0.4151* & 89.94* &	39.55* & 0.5460* & 92.07* & 52.69* & 0.6675* \\
        & MD & & 80.44 & \bf 15.31 & \bf 0.2439 & 90.17* &  \bf 31.05* &  \bf 0.4546* & 88.15*	&  \bf 43.84* & \bf 0.5822* & 92.22* & \bf 57.27* & \bf 0.7039* \\
        \hline
        \multirow{3}{5em}{\shortstack[c]{MVSNet~\cite{yao2018mvsnet} \\ softmin}} & Blended & \checkmark & 74.67	& 2.14 & 0.0395 & 83.66 & 11.68 & 0.1897  & 83.43 & 19.64 & 0.2974 & \bf 89.93 & 37.85	& 0.5240 \\
        & MD & \checkmark & \bf 84.60 & 11.70 & 0.1936  & \bf 88.65	& 30.99	&	0.4443  & \bf 85.93	& 43.16	& 0.5645 & 87.69 & 57.10 & 0.6870\\
        & MD & & 82.02 & \bf 14.71 & \bf 0.2371 & 83.67	& \bf 37.63 & \bf 0.5095 & 79.09 & \bf 51.40 &	\bf 0.6144 & 81.74 & \bf 63.66	& \bf 0.7129 \\
        \hline
        \multirow{3}{5em}{Vis-MVSNet~\cite{zhang2020visibility}} & Blended & \checkmark & 87.43 & 13.19 & 0.2158 & 87.98 & 35.82 &	0.4987 & 85.66 & 53.65	& 0.6541 & 89.26 & 64.99 & 0.7503\\
        & MD & \checkmark & \underline{\bf 91.44} & \bf \underline{21.72} & \underline{\bf 0.3330} & \underline{\bf  93.23} & \underline{\bf 48.22} & \underline{\bf 0.6305} & \bf 90.25 & \underline{\bf 64.40} & \bf 0.7485 & \bf 90.55	& \bf 72.88 & \bf 0.8067\\
        & MD & & 84.25 & 8.90 & 0.1491 & 84.19 & 32.45 & 0.4569 & 81.10 & 53.27 & 0.6352 & 86.13	& 68.03 & 0.7579 \\
        \hline
        \multicolumn{3}{c|}{COLMAP~\cite{schonberger2016pixelwise}} & 89.95 & 17.30 & 0.2785 & 93.20 & 42.76 & 0.5797 & \underline{94.88} & 62.68 & \underline{0.7511} & \underline{96.78} & \underline{73.64} & \underline{0.8346}
    \end{tabular}
\end{table*}

\section{Conclusion}

We have presented a study of the performance of deep MVS methods on Internet images, a setup that, to the best of our knowledge has never been explored. We discussed the influence of training data, architecture and supervision. For this last aspect, we introduced a new unsupervised approach which outperforms state of the art. Our analysis revealed several interesting observations, 
which we hope will encourage systematic evaluation of deep MVS approaches with challenging Internet images and stimulate research on unsupervised approaches, to ultimately have deep learning bring clear performance boosts in this important scenario.   

\subsection*{Acknowledgments}
\small
This work was supported in part by ANR project EnHerit ANR-17-CE23-0008 and was performed using HPC resources from GENCI–IDRIS 2020-AD011011756. We thank Tom Monnier, Michaël Ramamonjisoa and Vincent Lepetit for valuable feedback. 

{\small
\bibliographystyle{ieee_fullname}
\bibliography{egbib}
}

\end{document}